\documentclass[11pt,preprint]{article}

\usepackage[review]{acl}
\usepackage{textgreek}
\usepackage{times}
\usepackage{latexsym}
\usepackage{hyperref}
\usepackage[T1]{fontenc}
\usepackage{graphicx}
\usepackage{microtype}
\usepackage[english,bidi=default]{babel}
\usepackage{multirow}
\usepackage{multicol}
\usepackage{array}
\usepackage{here}
\usepackage{inconsolata}
\usepackage{footnote}
\makesavenoteenv{table*}
\usepackage{amsmath}
\usepackage{amssymb}
\usepackage[most]{tcolorbox}
\usepackage{xcolor}
\usepackage{tabularx}
\usepackage{booktabs}
\usepackage[table]{xcolor}

\definecolor{medscape}{HTML}{EAF2FB}
\definecolor{medicalxpress}{HTML}{F3EAFB}
\definecolor{framegray}{HTML}{4E5D6C}
\definecolor{accent}{HTML}{1F4E79}

\usepackage[T2A,T1]{fontenc}
\usepackage[utf8]{inputenc}

\newcommand{\bg}[1]{{\fontencoding{T2A}\selectfont #1\fontencoding{T1}\selectfont}}

\title{Does Machine Translation Retain Media Bias?}

\author{
  \textbf{Silvia V.\ Gargova}\textsuperscript{1} ~~~~~
  \textbf{Tim Menzner}\textsuperscript{2,3} ~~~~~
  \textbf{Jochen L.\ Leidner}\textsuperscript{2,3}\\[5mm]
  \textsuperscript{1} University of Alicante, Alicante, Spain \\
  \textsuperscript{2} Center for Responsible Artificial Intelligence (CRAI), Coburg, Germany\\
  \textsuperscript{3} Coburg University of Applied Sciences and Arts, Coburg, Germany
}

\date{}

\begin{document}

\maketitle

\begin{abstract}
  Within the most recent two decades, 
  machine translation (MT) has been advanced
  from a research area to a mature, deployed technology
  for prose text and most language pairs.
  However, language has different dimensions of
  meaning, and the alluded progress has 
  primarily been demonstrated for propositional
  meaning. 
  In this small exploratory study, we investigate 
  whether one type of affective meaning, namely
  different types of linguistically expressed media bias, are retained under automatic
  translation for two language pairs,
  English and German as the
  source languages, and Bulgarian as the 
  target language.
  We evaluate four state-of-the-art translation methods to answer the question
  whether and to what degree
  a loss of non-neutral language (specifically, 27 types of media bias) can occur as a result of applying MT.
  Our empirical findings are consistent with the
  claim that MT does not fully retain existing news
  bias of its input source text, but also may add
  extra bias of its own.
  To the best of our knowledge, this is the
  first experimental study of MT media bias
  retention.
  %
\end{abstract}

\section{Introduction}

\subsection{Background and Motivation}

The purpose of automatic machine translation is
to generate a text document in the target language
that is meaning-equivalent to the document in the
source document \cite{Koehn:2020}. While this can be challenging even for neutral,
fact-centric prose text, text is often biased:
\emph{media bias (also ``news bias") can be
defined as the conscious or unconscious
deviation from neutral (impartial) reporting}, e.g. by using
such rhetoric devices such as framing or word choice (viz.
``freedom flighter" versus ``terrorist").
In Leech's seven types of meaning system \cite{Leech:1981},
media bias belongs to the realm of \emph{associative
meaning}. How can machine translation system cope with preserving
the associative meaning dimension across languages?

To assess the exact position of a writer or news outlet
for a language one does not read, today, we may use automatic
machine translation. However, it is not clear whether
various forms of media bias are retained under automatic
translation or whether or to what extend such subtle information
literally gets ``lost in translation".

%
\textbf{What is the right outcome?}
One question in the course of answering this question
is what kind of behavior would users actually expect
from an MT system: to translate existing bias or to
take it out? We believe that the goal of MT should be
to carry over as many content and cultural nuances as possible, and that includes shades of associative meaning
such as news bias. The reader deserves to learn not only
about the propositional content but also about the
presence or absence of neutrality and the positioning
and opinions of the author of the text to be translated.

\subsection{Scope}

In this paper, we investigate this question using English and German as source languages and Bulgarian as the target language. To the best of our knowledge, this is the first study to investigate whether MT retains media bias across languages:

\begin{center}
\begin{tabular}{|l|}\hline
 \,\,\textbf{(RQ.)} \emph{ \,\,\, Does MT retain media bias? \,\,\, } \,\, \\ \hline
\end{tabular}\\
\end{center}

We present an exploratory study of news bias retention and introduction by MT. In addition, we introduce a first small, diagnostic corpus, BiasTrans-BG-1.0-small, which we make available to the research community\footnote{anon. repo link}.

\section{Related Work}

\subsection{Work on Automatic Machine Translation of Associative Meaning}

The aforementioned RQ is an instance of the
more general question to what extent MT
retains nuances of the various kinds of
associative meaning.

Past MT work has dealt with the effect of
MT on \emph{sarcasm} \cite{Sharma-Joshi:2023:ICSS},
\emph{irony} \cite{Zheng-etal:2024:NLPJ} and \emph{stance} towards \emph{rumours} \cite{scarton2021cross}, which is different in topic but perhaps closest in spirit to our work presented here. Furthermore, there is also a large body of research on the \emph{introduction} of \emph{gender bias} during the automatic translation of names of professions \cite{Stanovsky-etal:2019:ACL}, because some languages make the biological sex explicit in the grammatical gender of pronouns (e.g.\ \textit{The doctor ... she/he ...}) when in the source language the biological sex of the person referred to could remain unspecified.

We are not aware of any prior work on the
\emph{retention} of \emph{news bias}  during
machine translation.

\subsection{Work on Automatic News Bias Analysis.}
\citet{Recasens-etal:ACL} were the first to attempt to carry out automatic sentence-level  bias detection from prose text---see \citet{Hamborg-etal:2019:IJDL}
for a review of early work.
\citet{Lim-etal:2020:LREC} and
\citet{Wessel-etal:2023:SIGIR} deal with the annotation of media bias.
\citet{Lee-etal:2022:NAACL} present a
multi-document summarizer that aims to
resist framing bias.
\cite{DaSanMArtino-etal:2020:IJCAI}
surveys \citet{Spinde:2024}'s monograph represents the state of the art in automatic media bias detection as of late 2023.
\citet{Menzner-Leidner:2024:ECIR} 
describe a fine-tuned
neural transformer model \cite{Vaswani}, fine-tuned for detecting 27 types of news bias.

\subsection{Work on Media Bias for the Bulgarian Language}

Azizov and co-workers described
a method for cross-lingual bias-detection \cite{Azizov-etal:2024:FindEMNLP}, and they
considered Bulgarian as one of several
languages.

\section{Dataset}

We used a dataset from an existing news bias model deployed to the public, 
which permitted its users to donate stories that are interesting or that 
contain wrongly classified sentences for research purposes at the press
of a button.(Anonymized)
Besides these user submitted examples (the five most frequent sources being Tagesschau, Fox News, Bild, Die Zeit, and BBC) the dataset also includes samples from BABE\cite{spinde2021babe}, Reuters/Fox News\cite{menzner2025biasrank}, and TRC2/Reuters\cite{nist_trc2}, resulting in 2,196 unique sentences.

From this data, we selected a sample of $N=446$ sentences (2 subsets with 233 English
and 213 German sentences each) that were annotated with a multi-label schema (we only use isolated sentences, not full articles, to avoid copyright issues). We use the term ``sentence" for simplicity, in practice, what we call a sentence can also be a cohesive union of small snippets like ``massive landslide. No one can deny climate change anymore. Ban flights now!" -- Table \ref{tab:dataset} provides some summary statistics on the resulting dataset used for the experiments described in this paper.

Because our news bias model used was a single-class model, but sentences had
potentially multiple labels, we ran them through the news bias model and if the output label was also included in the human annotated set, we used it as the gold label; this effectively provided an automatic mapping from
a multi-class dataset to a single-class dataset, while ensuring the final label was human approved.

\begin{table}[t]
\centering
\caption{Summary Statistics of our Bias-Annotated Bilingual Corpus for the Translation Experiment.}
\label{tab:dataset}
\begin{small}
\begin{tabular}{|l|r|r|}\hline
\textbf{BiasTrans-BG 1.0} Dataset & DE-Subset & EN-Subset \\ \hline
Total number of characters      & 43,267 & 36,687 \\
Number of sentences             & 213    & 233    \\
Biased / unbiased sentences     & 102 / 111 & 141 / 92 \\
Mean sentence length (char.)    & 203.13 & 157.45 \\
Sentences w/ media bias (\%)     & 47.89  & 60.52  \\ \hline
\end{tabular}
\end{small}
\end{table}

\begin{figure}
    \centering
    \includegraphics[width=0.9\linewidth]{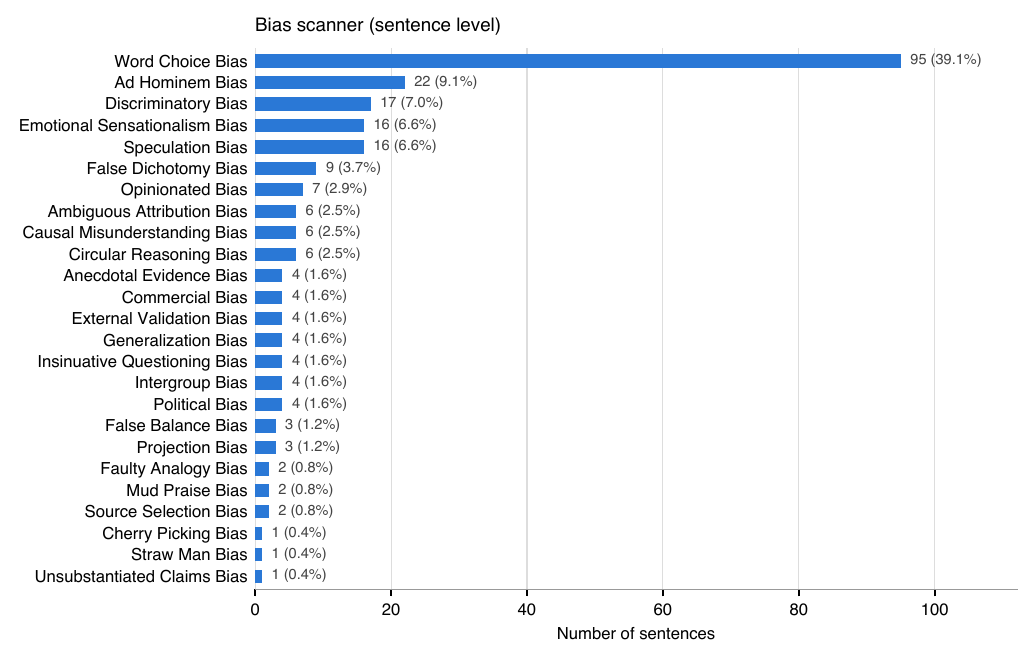}
    \caption{Histogram of Media Bias Types (across source languages)}
    \label{fig:hist}
\end{figure}
\section{Method}

Overall, our approach is the following:
First, all sentences of our gold dataset from English (EN) or German (DE) are translated into both Bulgarian (BG) and (the German ones also into) English (EN).\footnote{
                 BG will be used as the object language to answer
                our research question, whereas the EN translations may
                provide support to our human evaluator, a BG native
                speaker with no knowledge of DE but good command of
                EN.
             }
We elected to compare the machine translation capabilities of 4 automatic engines, namely:
DeepL's 
DeepL,\footnote{\url{https://www.deepl.com/de/translator}} 
Google 
Translate,\footnote{\url{https://translate.google.com/?sl=es&tl=pl&op=translate}} 
OpenAI 
GPT-5.5-2026-04-23,\footnote{\url{https://platform.openai.com/home}} and 
INSAIT BgGPT model \texttt{bggpt-gemma-3-27b-fp8} \cite{Alexandrov-etal:2024:ArXiv}.\footnote{\url{https://models.BgGPT.ai} (all accessed July 2026.)}
These were applied as Web services as of 
07/2026.
We used the
following prompt for LLM-based translations with GPT-5.5 and BgGPT:
\begin{quote}
    \small \texttt{\textbf{Role:} Expert translator. \\
        Translate the following text from \{language\} to Bulgarian. Return only the translation. \\
        Text: \{text\}
    }
\end{quote}
  Next, a news bias classification model must be applied to detect sentences that exhibit some form of news bias. We
        use the neural transformer model from BiasScanner \cite{Menzner-Leidner:2024:ECIR}, which performs well and
        can deal with many languages due to neural transformers' transfer
        learning ability. While it was trained and evaluated primarily on English, as part of a different study of ours, we also evaluated its quality on the Bulgarian language elsewhere (Anonymized during review, successfully completed peer review with publication TBD), where we found the Area under Curve (AUC) for distinguishing between biased and non-biased documents to be at 0.867 for Bulgarian, which is basically indistinguishable from the 0.865 scored on English. 
  
  Using his model, we conduct an automatic evaluation, i.e. essentially we check
        whether the automatic news bias model assigns the same category
        labels as the human annotator gold labels in our dataset. Where this is the case, for these
        sentences we can answer the research question in the affirmative.
  %
  In addition, we asked a human subject (and a native speaker of Bulgarian) to validate a sample of the sentences, both regarding the correctness of the translation in general and the bias label in specific. This step was added to avoid being prone to classifier noise or bias.

  Finally, we evaluated translation quality using the reference-free xCOMET metric\footnote{\url{https://huggingface.co/Unbabel/XCOMET-XL}}.\cite{Guerreiro-etal:2024:TACL} Overall, the translations achieved high scores ($\mu = 0.9139$, $M = 0.9632$), suggesting translations of high estimated quality and supporting their suitability for the bias classification task. DeepL achieved the highest mean score ($\mu = 0.9291$), followed by GPT-5.5 ($0.9162$), BgGPT ($0.9086$), and Google Translate ($0.9019$). English-to-Bulgarian translations also outperformed German-to-Bulgarian translations ($\mu = 0.9306$ vs.\ $0.8957$), while median scores across models remained high ($M \geq 0.9577$).

\section{Experimental Evaluation and Results}
\label{sec:results}


Then we compared the news bias classifications for $N=446$ paired source sentences.\footnote{We use the term \emph{sentence} for simplicity. In practice, a sentence in our dataset can also be a cohesive union of short snippets, such as ``Massive landslide. No one can deny climate change anymore. Ban flights now!''}
Each dataset sentence was evaluated before and after MT into Bulgarian
by BgGPT, DeepL, Google and GPT-5.5, and was used as one unit of analysis.
This resulted in 243 human-validated biased sentences and 203 validated unbiased sentences, for which both the human label union and the bias model output were empty.
The same dataset sentences were evaluated in five conditions (original plus 4 transl.). The
observations were therefore paired and could not be treated as independent
samples. 
We 
used a paired permutation test to compare the complete profiles of bias types. For one specific bias type, its presence or absence in a dataset sentence is a binary outcome. Cochran's $Q$ test \cite{Cochran:1950:Biometr} was used to compare such an outcome across more than two paired conditions, and the exact McNemar test \cite{McNemar:1947} was used for the subsequent pairwise comparisons. Because many bias types and systems were compared, the Holm correction \cite{Holm:1979:ScandJStat} was used to control the family-wise error rate.
We briefly summarize our findings.


\begin{figure}
    \centering
    \includegraphics[width=.85\columnwidth]{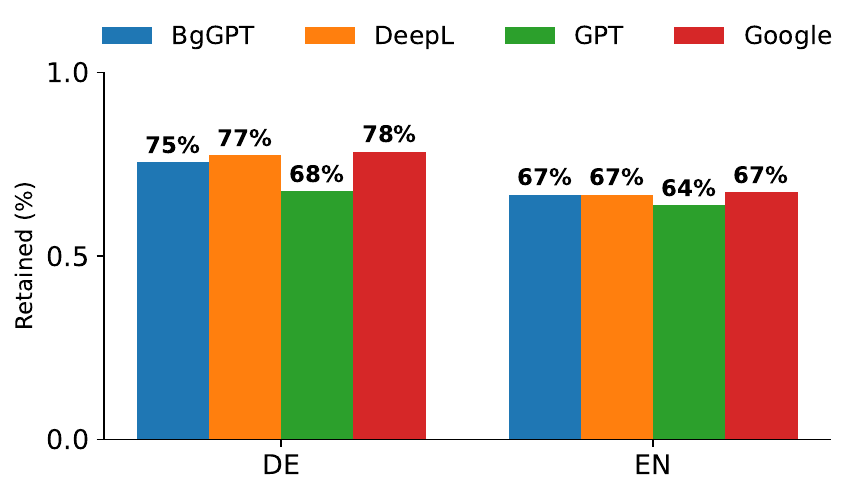}
    \caption{Retention of news bias gold labels after MT into BG, by source languages, and for four MT engines.}
    \label{fig:retention}
\end{figure}

\begin{figure}[H]
    \centering
    \includegraphics[width=\columnwidth]{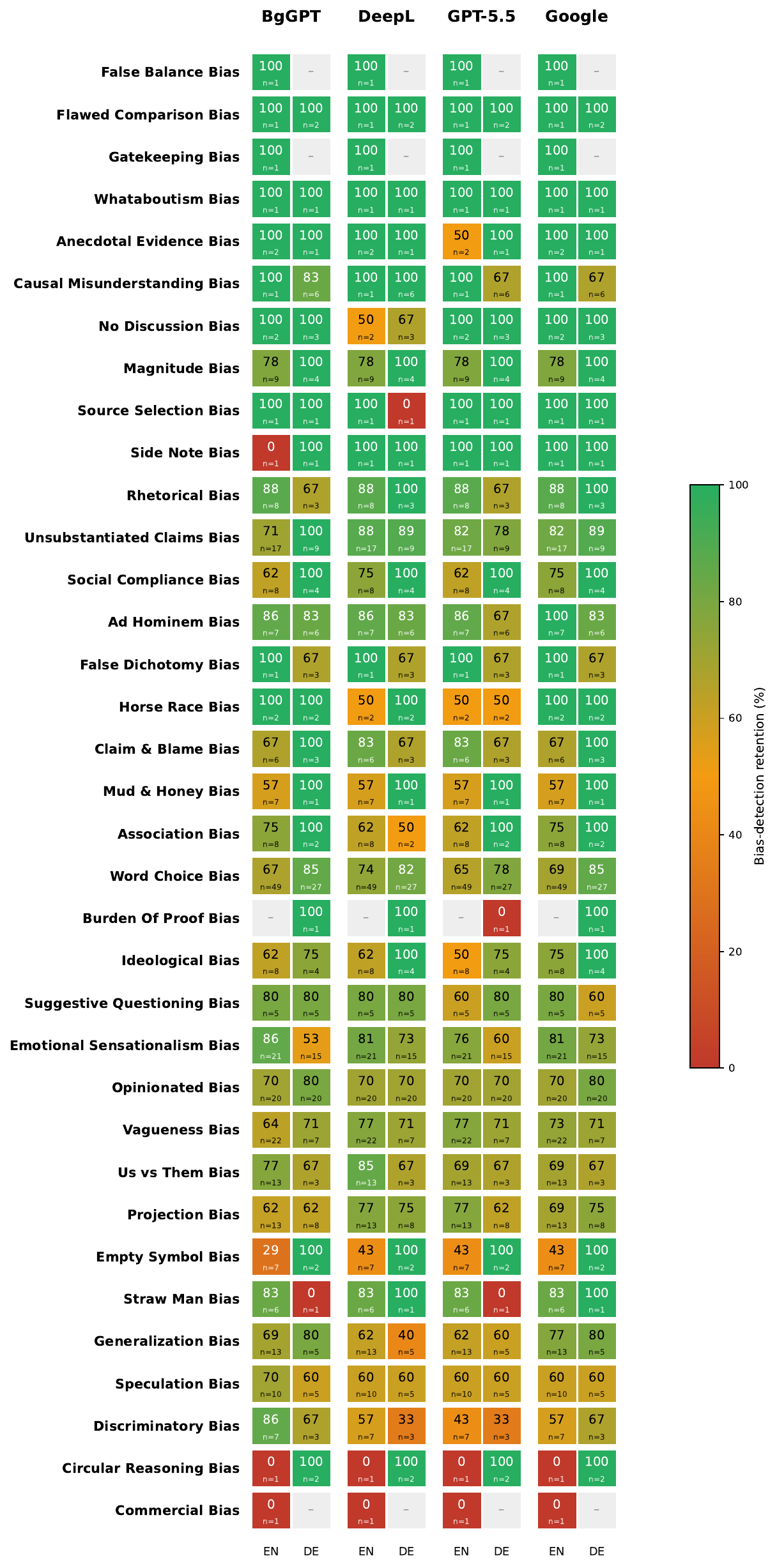}
    \caption{News bias (\%) after translation into Bulgarian, by category, source language, and MT engine. Each cell shows $n$ in second row; grey indicates no data.}
    \label{fig:heatmap}
\end{figure}
\subsection{Quantitative Evaluation}

Across all four translation methods, between 64\% and 78\% of news bias
is retained after a translation to BG (Figure \ref{fig:retention}). Looking at
it from the other end, between one third to one quarter of news bias gets lost in
the translation process. Retention is slightly higher for DE as a source language. compared to EN, despite LLMs almost certainly having seen more EN in pre-training, and inter-engine variability is clearly higher for DE than for EN.
%

%
%
The paired global permutation test indicated a significant difference among
the type profiles of all five conditions
($T_{\mathrm{sumQ}}=168.20$, $p_{\mathrm{MC}}<.001$). 

\begin{table*}
    \centering
    \small
    \caption{Qualitative Error Analysis: Example Sentences}
    \label{tab:error-analysis-example}
    \begin{tabularx}{\textwidth}{p{3cm}p{3cm}p{2cm}p{2cm}p{2cm}X}
        \toprule
        \textbf{Source text} & \textbf{Translated text} & \textbf{Gold label} & 
        \textbf{BiasScanner label (source text)} & 
        \textbf{BiasScanner label (translated text)} & 
        \textbf{Description} \\
        \midrule
        Die Industrie muss Gewinne erzielen, damit sie das Fliegen klimaneutral machen kann. &
        \bg{Индустрията трябва да реализира печалби, за да направи летенето климатично неутрално.} &
        Causal Misunderstanding Bias &
        Causal Misunderstanding Bias &
        --/-- &
        News Bias Not Retained (Lost in Translation: FN)\\ \hline
        
        The dissenting judges were from Russia, Cyprus and Romania. &
        \bg{Съдиите, които изразиха несъгласие, бяха от Русия, Кипър и Румъния.} &
         --/-- &
         --/-- &
        Discriminatory Bias &
        MT-introduced News Bias (FP)!\\ \hline
        
        Before we begin, remember this is coming from a man under investigation. &
        \bg{Преди да започнем, не забравяйте, че това идва от човек, който е разследван.} &
        Ad Hominem Bias &
        Ad Hominem Bias &
        Ad Hominem Bias &
        Retained  News Bias \\
        \bottomrule
    \end{tabularx}
\end{table*}
Holm-adjusted pairwise permutation tests showed significant differences between the original and each of the four translations (table omitted for space reasons, see (Anonymized)). 
None of the six pairwise comparisons among the Bulgarian translations was significant (all $p_{\mathrm{Holm}}=1.00$).
Thus, the global multi-label profile of the English original differed systematically from those of the BG translations, whereas no difference in the global profile was detected among the 4 translation engines.
%
%
%
%
%
%
%
%
%

\subsection{Qualitative Evaluation}

Table \ref{tab:error-analysis-example} shows a couple of English and German sentences and their Bulgarian MT equivalents, as well as the news bias labels before and after translation.

Figure \ref{fig:heatmap} reports, per bias category, source language, and translation engine, the share of gold-labelled biased excerpts an automatic scanner still detects as biased once translated into Bulgarian.

Retention depends on the kind of bias more than the engine (see Figure \ref{fig:heatmap}). Structural biases like omission, false balance, whataboutism, flawed comparisons survive translation almost perfectly (100\%), because their bias comes from what the story leaves out, not how it's phrased. Tone-dependent biases hold up less well: Circular Reasoning (50\%) and Discriminatory Bias (55\%) rely on nuanced wording that translation tends to smoooth. The two most frequent categories, Word Choice and Emotional Sensationalism are in the middle around 75\% - better than expected, suggesting MT often preserves loaded language rather than neutralizing it. 
The 0\% and 100\% extremes are almost all driven by categories with very few instances ($n\leq{}2$), so they are best treated as tentative (potential noise).
Observing beyond our original RQ as originally posed, the example in the second row indicates that \textbf{news bias can
even be introduced during MT}, i.e.\  our
line of experimentation proved more
impactful than anticipated.

\section{Summary, Conclusions and Future Work}

We investigated to what extent different types of media bias ``survive" going through
a machine translation process. For two language pairs, EN$\rightarrow$BG and DE$\rightarrow$BG,
we compared human-assigned class labels with machine-assigned
class labels after passing through a range of alternative,
state-of-the-art MT methods.

Our findings suggest that across the board, most translation engines retain between
64\% and 78\% of news bias in sentences; this means, at least medium to strongly biased news article will still be recognizable as such
even after undergoing automatic translation (e.g. as part of a recommender engine's preprocessing 
part of a news portal). To our surprise,
sometimes non-neutral reporting also gets
introduced during translation, which requires further investigation, and which
calls for extending our research question to
"To what extent does MT \underline{introduce or} retain
media bias?".

In future work, further language pairs and additional MT
systems should also be investigated. 
MT engines should be investigated  with respect to the retention of non-propositional semantic nuances
other than news bias.
%


%
%



\section{Limitations}

Despite experimenting with data covering several hundred sentences in two language pairs and evaluating three dedicated translation models and one foundational model instructed to translate by a prompt, there remain a number of limitations: importantly, in this study, we cold not control all variables: two language pairs are not \emph{all} language pairs -- there
exist thousands of languages and $k(k-1)$ pairs of them. There are also more than $K=4$ models
or engines that can be used to translate text.
We used only one news bias models in our experiments; again, one could try more and
different models.

\begin{figure}
    \centering
    \includegraphics[width=\linewidth]{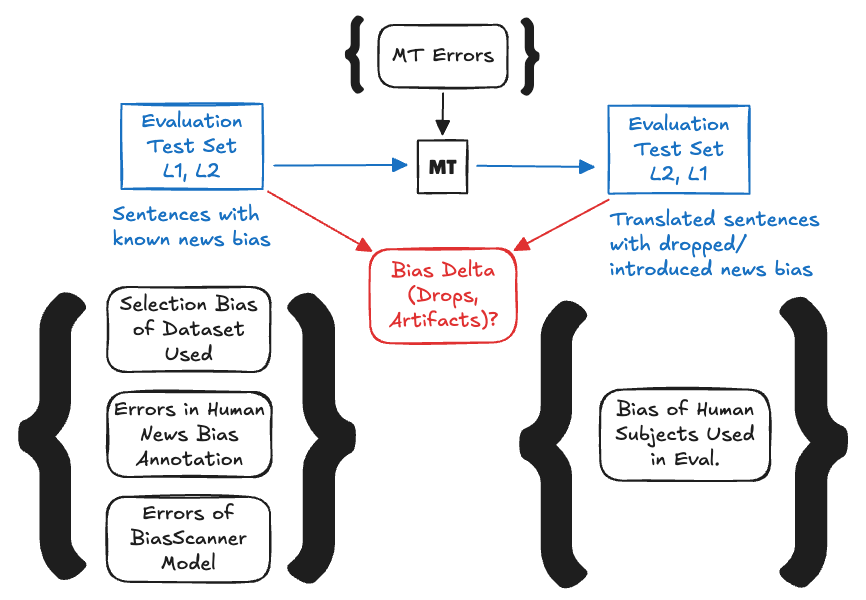}
    \caption{Experimental Design}
    \label{fig:experimental-design}
\end{figure}

In our task´s experimental design, ideally the following factors
need to be considered and controlled as well as possible (Figure \ref{fig:experimental-design}).
\begin{itemize}
\item \textbf{Selection News Bias of Source Document Used in Experiment.}
It is not clear what would be an adequate size of a corpus
for an evaluation experiment the validity of which can be easily defended, but it would certainly have to be larger than
ours; clear is that we want it to contain sufficient
biased and unbiased sentences for all language pairs under
consideration, and that it cover both directions ($L_1\rightarrow{}L_2$ and $L_2\rightarrow{}L_1$). Harder is the question what it would contain, i.e. how
to sample its constituent sentences. 

\item \textbf{Annotation Errors of Source Document´s News Bias Annotation.}
The source language sentences have been annotated for news
bias, and the labels can be wrong or at least biased.

\item \textbf{Errors of the BiasScanner Model.}
The errors of BiasScanner´s model have been evaluated by its
authors, but no specific score has been published for
Bulgarian language input.

\item \textbf{Bias of Human Subjects Used to Evaluate.}
It is usually accepted that human subjects add their own
issues to an experiment, and the standard way to mitigate
them is to use several of them, and to report inter-annotator
agreement. In this small exploratory our only available speaker
of Bulgarian was author one; a larger study should use e.g.
three or five subjects and report
Krippendorff´s $\alpha$ \cite{Krippendorff:2004,Krippendorff:2004:HumCommRes}.
\item \textbf{Machine Translation Quality Errors.} The
MT systems' quality is relevant and it should be
established; quality issues are adversely influencing
what we measure, and they are at the same time part of
what we want to measure here: dropping an instance of bias or introducing a bias artifact during MT are
both examples of quality problems. We checked with xCOMET
and established high MT quality across all engines used in this study.
\end{itemize}
Because of this multitude of potentially confounding forces and noise sources at play and the limited scope of this short paper, we cannot legitimately make a strong claim like (in general) "MT does not fully retain bias." or "MT causes bias." Nevertheless, we \emph{can}
claim that \textbf{our findings are consistent with both these claims}.

We believe that many or most of our observations
obtained would be confirmed on larger samples, more language pairs, and/or additional news bias models, even though the shape of the bias histogram before and after translation depends
on the nature of the news bias model and the
exact nature of the translation engine, and we would
like to encourage the research community to explore this
topic neglected to date.

\section{Ethical Considerations}

\textbf{Dual Use.} Our study of the effect of machine translation on news bias retention is intended to answer a research question to enrich the body of scientific knowledge, an ethically legitimate goal. 

Nevertheless, our study can be viewed as a \emph{dual good} in the sense that a morally perturbed party could also use our method or data to study concealing or adding news bias by machine. We believe in open science for good with no censorship, in according to an open society.

To mitigate any threats, we require organizational registration before  users can download our data.

\textbf{Data Collection and Annotation by Human Subjects.} Co-authors 1 and 2 contributed to the background (BG) review and bias annotations, respectively. Sentences selected for annotation were sampled from a dataset volunteered by anonymous web users, who acted under informed consent (we applied for a waiver
letter from our local IRB, as the 
ethics approval process is only defined
for \emph{named} individuals).

\textbf{AI model Use.} 
We used AI models for code bug fixing assistance, 
spelling and style improvements and \LaTeX{}-formatting help.

\section*{Acknowledgments.} We would like to gratefully
acknowledge the valuable feedback received by three
anonymous reviewers, whose detailed reading vastly increased the quality of this paper. All remaining
errors are of course ours.

The authors would like to acknowledge the funding
under the European ERASMUS+ programme (grant
2025-1-DE01-KA131-HED-000325540 awarded to the third
author) 
that made this collaboration possible.
We would also like to thank Veronika Wagner for helping to navigate the application process.



\bibliography{Gargova-Menzner-Leidner-2026-ArXiv}

\end{document}